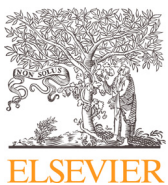
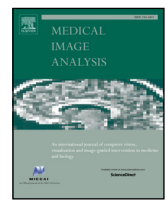

# Nonrigid reconstruction of 3D breast surfaces with a low-cost RGBD camera for surgical planning and aesthetic evaluation

R.M. Lacher[a], F. Vasconcelos[a], N.R. Williams[c], G. Rindermann[d], J. Hipwell[b], D. Hawkes[a], D. Stoyanov[a],*

[a] *Wellcome / EPSRC Centre for Interventional and Surgical Sciences (WEISS), University College London, London, UK*
[b] *Centre for Medical Image Computing (CMIC), University College London, London, United Kingdom*
[c] *Surgical & Interventional Trials Unit, University College London, London, United Kingdom*
[d] *Independent Researcher, Stuttgart, Germany*

**ABSTRACT**

Accounting for 26% of all new cancer cases worldwide, breast cancer remains the most common form of cancer in women. Although early breast cancer has a favourable long-term prognosis, roughly a third of patients suffer from a suboptimal aesthetic outcome despite breast conserving cancer treatment. Clinical-quality 3D modelling of the breast surface therefore assumes an increasingly important role in advancing treatment planning, prediction and evaluation of breast cosmesis. Yet, existing 3D torso scanners are expensive and either infrastructure-heavy or subject to motion artefacts. In this paper we employ a single consumer-grade RGBD camera with an ICP-based registration approach to jointly align all points from a sequence of depth images non-rigidly. Subtle body deformation due to postural sway and respiration is successfully mitigated leading to a higher geometric accuracy through regularised locally affine transformations. We present results from 6 clinical cases where our method compares well with the gold standard and outperforms a previous approach. We show that our method produces better reconstructions qualitatively by visual assessment and quantitatively by consistently obtaining lower landmark error scores and yielding more accurate breast volume estimates.

## 1. Introduction

Breast cancer is the most frequently diagnosed cancer site among women worldwide (Jemal et al., 2011; Fitzmaurice et al., 2015). Despite increased incidence, mortality from breast cancer is declining with 10-year survival rates reaching 82% in Europe (De Angelis et al., 2014). A longer life expectancy after developing breast carcinoma in turn emphasizes the importance of aesthetic treatment outcome and late effects. Beside the oncological result, several studies have linked cosmetic and functional outcome to patients' quality of life, mental health and self-image (Hau et al., 2013; Stanton et al., 2001). A good cosmetic outcome is typically associated with high breast symmetry and minimal scarring. In breast conserving surgery also known as lumpectomy the tumour with clear tumour-free margins is excised usually followed up by local radiation of the treated breast therefore allowing the patient to keep most of her breast in the setting of early stage breast cancer. This is backed by large long-term clinical studies concluding no difference in disease-free survival between lumpectomy and mastectomy (Fisher et al., 2002). While breast conserving surgery generally surpasses mastectomy regarding their cosmetic outcome, dissatisfactory or poor results are reported for lumpectomy in up to 30% and 6% of cases respectively (Hill-Kayser et al., 2012). Yet, a lack of standardised procedures for aesthetic outcome evaluation persists (Cardoso et al., 2012). For a higher patient satisfaction and fewer adverse cosmetic results it is essential to correlate tumour and treatment related factors to breast aesthetics post-treatment as well as further involve patients in the decision making process between the rising number of therapeutic alternatives. 3D surface imaging of the breast has the potential to aid in treatment planning, surgical outcome prediction and objective aesthetic outcome evaluation (Chae et al., 2016; O'Connell et al., 2015; Cardoso et al., 2014). Computational tools start to emerge that incorporate 3D breast surface data to let clinicians appreciate real-istic 3D visualisations of the breasts' morphology on the click of a button, perform shape-related analysis and classification as well as biomechanical simulation of probable surgical outcomes overall showing their utility to replace and extend former crude time-consuming and subjective techniques (Eiben et al., 2016; Oliveira et al., 2014). Unfortunately, a widespread clinical use is elusive due to the high cost and infrastructure requirements of 3D scanner equipment (Tzou et al., 2014). We therefore investigate the use of a portable consumer-grade RGBD camera for complete accurate 3D breast surface modelling from breast data acquired in a simple contactless acquisition setup in hospital. In this paper in particular, we address the handling of postural sway, the nonrigid involuntary body deformation and breathing motion during incremental data acquisition, and demonstrate our method's superiority in terms of reconstruction quality compared to a previous method.

* Corresponding author.
*E-mail addresses:* rene.lacher.13@ucl.ac.uk (R.M. Lacher), f.vasconcelos@ucl.ac.uk (F. Vasconcelos), n.williams@ucl.ac.uk (N.R. Williams), gerrit.rindermann@gmail.com (G. Rindermann), j.hipwell@ucl.ac.uk (J. Hipwell), d.hawkes@ucl.ac.uk (D. Hawkes), danail.stoyanov@ucl.ac.uk (D. Stoyanov).

## 2. Previous work

3D surface reconstruction models the external geometry and appearance of real world scenes. To recover a complete surface model of a physical object through perspective projection, data from several overlapping images has to be merged. This requires knowledge of the camera intrinsic parameters and egomotion. With recent ubiquity of cheap portable RGBD cameras delivering medium resolution depth at video frame rate there is a vast literature on image-based 3D surface reconstruction spanning ample application areas in healthcare and beyond. Static scene reconstruction has reached maturity in recent years (Newcombe et al., 2011). State-of-the-art methods are typically posed as a Simultaneous Localisation and Mapping (SLAM) problem employing a variant of Iterative Closest Point (ICP) (Besl et al., 1992) for frame-to-model tracking and an incremental volumetric (Curless and Levoy, 1996) or point-based fusion (Alexa et al., 2003; Pfister et al., 2000). Although follow-up research addressed scalability (Nießner et al., 2013; Whelan et al., 2015), robustness (Dou et al., 2016; Zhou and Koltun, 2014; Whelan et al., 0000; 2015; Keller et al., 2013; Glocker et al., 2015) and global consistency (Henry et al., 2012; Maier et al., 2014; Dai et al., 2017; Zhou and Koltun, 2013; Whelan et al., 0000) the assumption of a rigid camera motion remains restrictive.

In practice, real world scenes frequently comprise parts deforming to varying degrees necessitating a solution to the more challenging problem of nonrigid reconstruction. The latest work on nonrigid reconstruction of human subjects with a single Kinect-style camera continues to rely on closest point correspondences in the form of projective data association (Newcombe et al., 2015; Innmann et al., 2016), normal shooting (Li et al., 2013) or probabilistic models (Cui et al., 2012) only supporting limited inter-frame motion and and little change in surface topology. Criteria to prune correspondences usually include distance thresholds (Li et al., 2013; Zollhöfer et al., 2014; Innmann et al., 2016), a normal orientation incompatibility check (Zollhöfer et al., 2014; Zeng et al., 2013; Li et al., 2013), visibility or boundary constraints (Zollhöfer et al., 2014; Newcombe et al., 2015; Li et al., 2013; Zeng et al., 2013) or bijective correspondences (Zeng et al., 2013). Volume grids of truncated signed distance function values remain popular in nonrigid reconstruction in its function as a rigid reference space (Newcombe et al., 2015; Innmann et al., 2016) along point-based data structures (Li et al., 2013; Cui et al., 2012; Zeng et al., 2013) or template tracking (Zollhöfer et al., 2014). Some methods introduce a separate surface detail integration step (Li et al., 2013; Zollhöfer et al., 2014). Most commonly a two-staged registration strategy estimates initial rigid camera poses followed by a nonrigid deformation (Innmann et al., 2016; Zollhöfer et al., 2014; Li et al., 2013; Newcombe et al., 2015; Zeng et al., 2013). Correspondence error is measured as a point-to-point distance (Bogo et al., 2015; Amberg et al., 2007), point-to-plane distance (Newcombe et al., 2015; Dou et al., 2016; Innmann et al., 2016) or a combination of both metrics (Zollhöfer et al., 2014; Zeng et al., 2013; Li et al., 2013; 2009) usually embedded in a robust kernel (Zollhöfer et al., 2014; Newcombe et al., 2015; Li et al., 2013; Bogo et al., 2015). Occasionally a dense colour term penalises RGB discrepancy (Zollhöfer et al., 2014; Bogo et al., 2015) or sparse colour feature constraints are used (Innmann et al., 2016). Frames are aligned in a frame-to-model fashion (Newcombe et al., 2015; Zeng et al., 2013) or jointly (Li et al., 2013). Assuming small motion between consecutive frames, most methods pursue an as-rigid-as-possible approach imposing local smoothness of the deformation (Zeng et al., 2013; Zollhöfer et al., 2014; Li et al., 2013; Newcombe et al., 2015; Bogo et al., 2015) and enforcing isometry (Innmann et al., 2016; Zollhöfer et al., 2014). Local deformations are either rigid (Newcombe et al., 2015; Innmann et al., 2016) or affine (Li et al., 2013; Zollhöfer et al., 2014). Methods can further be divided into extrinsic deformation models that remap space through discrete volumetric warp fields that enclose the object (Zollhöfer et al., 2014; Newcombe et al., 2015; Innmann et al., 2016) and intrinsic deformation models such as embedded graphs that induce localised deformation on nearby space (Zeng et al., 2013; Li et al., 2013). Both stand in contrast to parametric subspace models using shape models (Bogo et al., 2015; Weiss et al., 2011) or articulated motion (Cui et al., 2012). In optimising the composite cost function, most methods employ a numerical nonlinear solver (Zollhöfer et al., 2014; Li et al., 2013; Newcombe et al., 2015), GPU acceleration (Zollhöfer et al., 2014; Newcombe et al., 2015; Innmann et al., 2016) and sparsity pattern exploits (Li et al., 2013; Zollhöfer et al., 2014). Convergence is improved through progressive relaxation of rigidity (Zollhöfer et al., 2014; Li et al., 2013; Amberg et al., 2007) or coarse-to-fine schemes applied to grid sampling (Newcombe et al., 2015; Innmann et al., 2016), template resolution (Zollhöfer et al., 2014) or shape model complexity (Bogo et al., 2015). To generate a final mesh model volumetric techniques commonly run marching cubes (Lorensen and Cline, 1987) while point-based methods depend on Poisson surface reconstruction (Kazhdan et al., 2006).

There is less literature on 3D breast surface imaging in particular nonrigid alignment of breast surfaces. Commercial single-shot 3D scanners find limited use in clinical practice despite a superior reconstruction quality due to their generally high cost and space requirements (Chae et al., 2016). Studies reviewing the Kinect's performance for breast surface imaging and other healthcare applications found its accuracy suitable for clinical use (Pöhlmann et al., 2016; Choppin et al., 2016) yet pointing out a large uncertainty in the measured values (Choppin et al., 2016) and a need for further clinical validation studies (O'Connell et al., 2015). A drawback of more affordable and portable 3D scanners are longer capture times, which inevitably lead to postural sway and thus lower the reconstruction quality. Postural sway is the entirety of involuntary body deformation and breathing motions between consecutive breast images in a series. 'Hold breath' scanning protocols and respiratory gating techniques have been proposed to counter its effects (Patete et al., 2013). Among studies using single or two Kinect camera setups for breast surface imaging (Oliveira et al., 2014; Wheat et al., 2014; Costa et al., 2014; Pöhlmann et al., 2014; Lacher et al., 2015) only one produces complete denoised 3D breast surface models from a sequence of depth images (Lacher et al., 2015). Although discussing its apparent negative impact on reconstruction quality the work lacked a way of handling postural sway entirely. In summary, the specific contributions of our work are:

- A template-free nonrigid reconstruction pipeline running only on a standard PC taking a noisy RGBD input video from a Kinect-style camera. We demonstrate the method's ability to create clinical-quality 3D breast surface models in spite of varying degrees of postural sway during data capture.

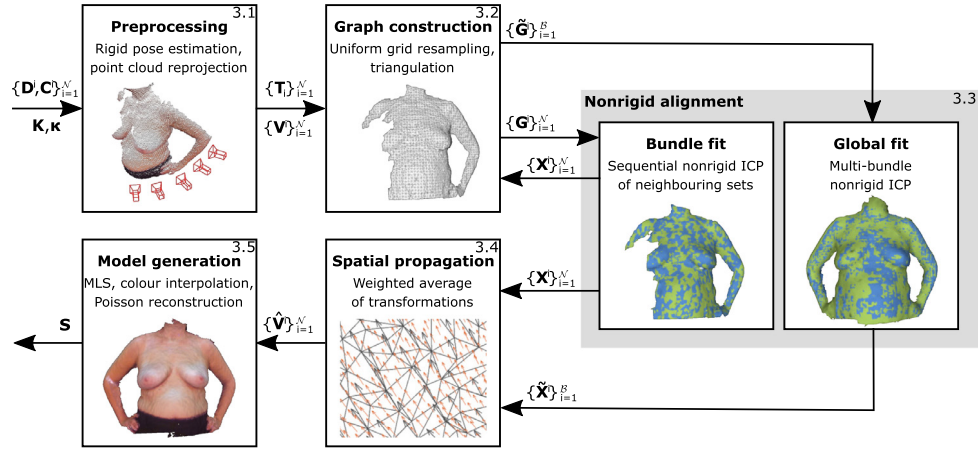

**Fig. 1.** Method overview. A sequential SLAM method (Lacher et al., 2015) is used to rigidly estimate camera poses (3.1). To deal with the vast number of transformation parameters lower resolution meshes are constructed by uniform downsampling and triangulation constituting a deformation graph structure as described in Sumner et al. (2007) (3.2). The pairwise nonrigid ICP method proposed in Amberg et al. (2007) is utilised to refine the alignment of all frames in the data set nonrigidly (3.3). Refinement is organised in two phases: a bundle (local) and a global phase. Resulting node-wise affine transformations are subsequently interpolated to deform the full resolution point clouds (3.4) prior to fusion and meshing (3.5).

- An extension of pairwise nonrigid ICP to the multiview case incorporating soft mobility constraints in areas of non-overlap.
- The proposition of shortest distance correspondences as a new technique for data association. We match source and target through finding the shortest intersecting line from any one source vertex with the target model. We show that shortest distance correspondences consistently lead to better alignment.
- A two-fold landmark and breast volumetric quantitative validation in metric units demonstrating improved reconstruction quality towards gold standard level and superior to a competing method.

## 3. Method

We propose a markerless template-free nonrigid reconstruction method for accurate 3D breast surface modelling in breast cancer treatment planning and evaluation in the presence of postural sway (see Fig. 1). Our method is based on the works of Amberg et al. (2007) and Sumner et al. (2007) combining the linear problem formulation for pairwise nonrigid alignment of the former with the notion of an embedded deformation graph from the latter. We use bold upper and lower case Latin letters for matrices and vectors. The elements of either are enclosed in square brackets. Scalars are written in lower case Latin and Greek letters with normal font weight. Capital Greek and calligraphy letters denote sets and constants respectively.

### 3.1. Preprocessing

The preprocessing stage includes intrinsic and extrinsic camera calibration for undistortion, RGBD registration and point cloud backprojection. In addition, an erosion of an automatically segmented foreground mask excludes spurious points with unreliable colour attribution.

#### 3.1.1. Calibration

Planar checkerboard images were acquired to calibrate the Kinect's internal infrared and RGB camera as projective pinhole cameras (Hartley and Zisserman, 2000). The calibration procedure estimated the camera intrinsic matrices $K_D$ and $K_{RGB}$ which define a linear mapping from metric to pixel space. Radial lens distortion was modelled as a polynomial function through a set of distortion coefficients (Zhang, 1999). The inverse distortion of pixel coordinates was computed in a Levenberg-Marquardt optimisation. Stereo calibration recovered the camera extrinsic matrix $T_{D\rightarrow RGB} = \{R_{D\rightarrow RGB}, t_{D\rightarrow RGB} \mid R_{D\rightarrow RGB} \in SO(3), t_{D\rightarrow RGB} \in \mathbb{R}^3\}$ which is a rigid transformation between the infrared and RGB sensor. Reprojection errors were below 0.2 pixels.

By means of timestamping, depth and colour images are paired. Following undistortion of pixel coordinates, depth values are back-projected into metric space producing an unordered cloud of 3D points in depth camera coordinates using the previously estimated camera intrinsics $K_D$. Invalid depth measurements, e.g. in occlusion areas between the infrared sensor and pattern projector, are skipped. After a change of coordinate system from depth to colour camera 3D point coordinates are reprojected to colour camera pixel coordinates for RGB value lookup. Each 3D point is assigned an RGB colour subject to nearest integer interpolation and image boundary checks. To reduce the computational cost of repeated point cloud creation, the RGB values are permanently stored in a new depth-registered colour frame. Fig. 2 illustrates the projection of depth data on the colour image for point cloud extraction.

#### 3.1.2. Foreground mask erosion

The Kinect camera has been reported to suffer from high noise levels in particular at depth discontinuities (Sarbolandi et al., 2015). Adding to this, discretisation in conjunction with calibration inaccuracies leads to gross errors at depth boundaries causing e.g. background-blending artefacts due to the projective nature of the camera transformations. While the reconstruction method is capable of smoothing out noise, fusing a large number of frames into a single representation, extra caution is required handling regions of inhomogeneous depth. In our controlled acquisition, depth discontinuities mostly occur between the foreground subject and background. We segment the foreground in each depth frame by determining the distance to the background wall. We calculate the median of the 9-by-9 right top corner image patch, obtaining a binary foreground mask $F(u) : \mathbb{N}_0^2 \mapsto [0.1]$ with $u \in \mathbb{N}_0^2$ denoting the integer grid pixel coordinates. This is followed up by 3 binary erosions $F \ominus O$ of the foreground mask with a rectangular structuring element $O$ of size $3 \times 3$ of all ones.

#### 3.1.3. Rigid pose estimation

We employ a sequential visual SLAM for breast surface reconstruction (Lacher et al., 2015) to estimate global camera poses $\{T_i\}_{i=1}^{\mathcal{N}}$ with $\mathcal{N}$ being the number of frame pairs in the RGBD se-quence and $T_i = \{R_i, t_i\}$ the individual rigid body camera transformation comprising a rotation matrix $R \in SO(3)$ and a translation vector $t \in \mathbb{R}^3$. The SLAM is based on the popular KinectFusion algorithm (Newcombe et al., 2011) fusing consecutive frames with small inter-frame motion in a discrete truncated signed distance function volume after frame-to-model ICP alignment under point-to-plane error metric. The camera poses initialise our nonrigid alignment algorithm by rigid transformation of the point data into a shared global 3D coordinate system.

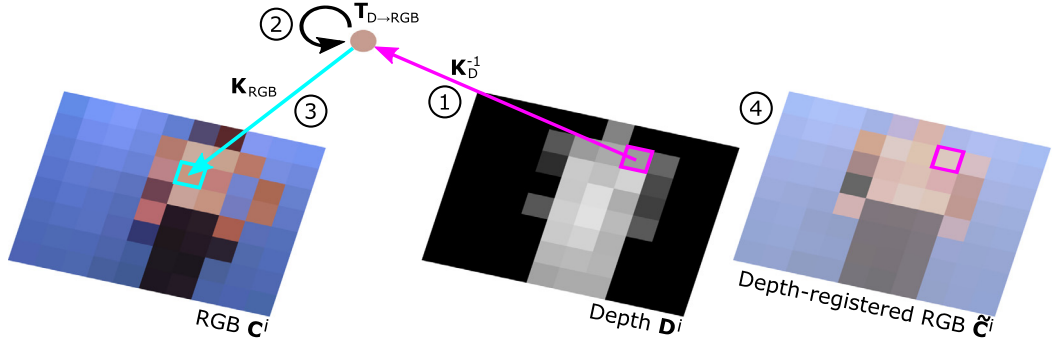

**Fig. 2.** Illustration of RGBD registration. Depth values are backprojected to 3D points using the inverse of the depth camera intrinsic matrix (1). After change of coordinate system (2) 3D points are projected to colour camera pixel coordinates for RGB value lookup (3). For convenience of point cloud creation RGB values are stored in a depth-registered colour image (4).

### 3.2. Graph construction

Each 3D point is assigned a normal $n \in \mathbb{R}^3$. Normals are computed in a pointwise parallelisation on the GPU. The normal orientation is estimated through principal component analysis of all 3D points in a projective depth window of size 3. The ambiguity in normal orientation is resolved by flipping normals consistently towards the origin of the camera coordinate system. This is achieved trivially by enforcing the dot product between the vector of the point coordinates (viewpoint is at origin) and the normal to be consistently negative or positive. Sumner et al. (2007) suggest the use of a coarse embedded graph to recover realistic shape deformations at human scale. We follow suit by downsampling all source point clouds via voxel grid averaging to a uniform sampling density of 6 mm. The downsampled point clouds are meshed employing the greedy projective triangulation algorithm with a search radius of 2.5 cm and a $\mu$ of 2.5 (Marton et al., 2009). Following triangulation, we perform a connected component analysis by iteratively traversing the triangular mesh. We discard smaller unconnected parts that cause a rank deficiency in the coefficient matrix. The triangulated downsampled single component point clouds serve as our deformation graphs $\{G^j\}_{j=1}^{\mathcal{N}}$ with $G^j = \{g_i^j\}_{i=1}^{\mathcal{J}} \subset \mathbb{R}^3$. $\mathcal{J}$ denotes the number of nodes in $G^j$. Each graph node $g_j^i$ is related to an affine transformation specified by a matrix $X_i^j \in \mathbb{R}^{4 \times 3}$.

### 3.3. Nonrigid ICP

Globally consistent nonrigid alignment is achieved by exhaustive joint alignment of all deformation graphs. For computational efficiency, each 10 consecutive frames are pooled into bundle graphs $\{\tilde{G}^i\}_{i=1}^{\mathcal{B}}$ where $\mathcal{B} = \mathcal{N}/10$ denotes the number of bundles. Each bundle is nonrigidly aligned by sequentially registering all other 9 deformation graphs in the bundle to the first of each bundle producing transformations $\{X^i\}_{i=j}^k$ with $k - j = 10$. Each bundle graph is then simultaneously aligned against the joint set of all other bundle deformation graphs. Prior to alignment, the joint sets of bundle graphs are resampled to ensure uniform node density. The alignment counts as a step and in turn consists of a maximum of 10 nonrigid ICP iterations. The alignment process iterates over all bundles in a circular fashion terminating after a maximum number of 100 steps has been reached. Each step, bundle transformation matrices $\{\tilde{X}^i\}_{i=1}^{\mathcal{B}}$ are incrementally updated.

#### 3.3.1. Correspondence finding

Correspondences from the downsampled source to target graph are sought on the grounds of spatial proximity. We compare two correspondence estimation techniques. In addition to classic closest point correspondences we investigate shortest distance correspondences. In the latter case, the correspondence is established using the intersection point at the end of the shortest line connecting the source vertex with the target mesh. The closest point correspondence search is sped up using an octree structure initially populated with the target's vertices. We implement the algorithm by Jones (1995) on the GPU for fast intersection point computation. The closest intersection point may either coincide with a target vertex, edge or lay within a triangular face. We omit superscript indices for readability. The projection of a source vertex $g_i$ on the plane of the respective target triangular face denoted as $g_i'$ simplifies to the following expression

$$g_i' = g_i - \langle g_i - f_1, n_f \rangle \cdot n_f \tag{1}$$

where $n_f$ is the unit face normal and $f_1$ an arbitrarily chosen face corner. In case $g_i'$ falls within the triangle it is also the intersection point $g_i''$. Assuming $g_i'$ lies outside the triangle and is instead closer to edge $e_{1,2} = f_2 - f_1$, we find the intersection point $g_i''$ on the edge as follows.

$$g''_i = g_i' + \frac{\langle f_1 - g_i', r \rangle \cdot r}{\langle r, r \rangle} \quad \text{with} \quad r = [(f_2 - g_i') \times (f_1 - g_i')] \times e_{1,2} \tag{2}$$

A complete geometrical derivation is found in Jones (1995). Per iteration the set of source graph nodes are updated by multiplication with the current matrix $X$. As illustrated in Fig. 3 and in accordance with Amberg et al. (2007), correspondences are clipped if one or more of the following conditions apply.

$$w_i \neq 0 \text{ iff } \begin{cases} d_{corr} < 0.02\,m & \text{and} \\ g_i'' \notin \Psi \subset \mathbb{R}^3 & \text{and} \\ \cos^{-1}(\langle n_i, n'' \rangle) < 45° \end{cases} \tag{3}$$

The normals $n_i$ and $n''$ denote the normals of $g_i$ and $g_i''$ respectively. Given the target is a non-watertight surface, $\Psi$ denotes the set of edges that make up the target's border. The border edges are the set of all single edges. A single edge is an edge that is only part of one triangle. Correspondences that are rejected carry zero weight. The weight for accepted correspondences is set to constant

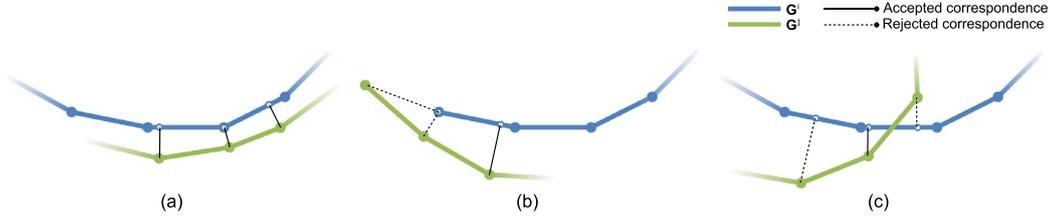

**Fig. 3.** Illustration of correspondence estimation and rejection for nonrigid alignment by the example of 2D contours. a) For every source vertex a closest intersection point with the target is determined to yield preliminary correspondences. These exact correspondences are superior to nearest point correspondences especially with coarsely resolved models. Correspondences are subsequently trimmed. b) Source vertices in a non-overlap region mapping to target vertices or edges belonging to the target's boundary are discarded as a correspondence and instead softly enforced to stay in place. c) Correspondences that exceed a distance threshold (leftmost correspondence) or whose normals deviate strongly (rightmost correspondence) are equally rejected. For normal comparison, the normal at the target intersection point is barycentrically interpolated.

1. The first two criteria ensure the algorithm is capable of dealing with non-overlap regions correctly while the last criterion prevents false correspondences when the still misaligned source and target intersect in regions where several geometric layers are close to each other. In the case of the torso, the latter may happen at the armpit or where the fingers are placed on the hip.

### 3.3.2. Data term

The data term $\epsilon_d$ is expressed as the sum of weighted quadratic point-to-point distances of correspondences. For simplicity of notation we assume the corresponding matching point in graph k for the query point $g_i^j$ in graph j to be $g_i^k$. As outlined in Section 3.3.1, $g_i^k$ is the nearest intersecting point with the target graph and need not be an actual node in the target graph $G^k$. It follows that $\Theta = \{(g_1^j, g_1^k), .., (g_{\mathcal{J}}^j, g_{\mathcal{J}}^k)\}$ is the set of fixed correspondences between the graphs j and k for a particular step and iteration.

$$\epsilon_d := \sum_{g_i^j \in G^j} w_i ||X_i \cdot g_i^j - g_i^k|| \quad (4)$$

Eq. (4) may be vectorised and rearranged to be amenable to linear solving through post-multiplication with $X$. Let $W$ constitute a $\mathcal{J} \times \mathcal{J}$ diagonal weight matrix with $diag(W) = [w_1, .., w_{\mathcal{J}}]$. Let $\hat{G}^j \in \mathbb{R}^{\mathcal{J} \times 4\mathcal{J}}$ be a matrix containing the graph's formatted nodes. Further we define $G$ to store the graph's nodes in a row-major matrix as written in Eq. (5).

$$\epsilon_{data}(X) := ||W(\hat{G}^j X - G^k)||_F^2$$

$$= \left\| \begin{bmatrix} w_1 & & & \\ & w_2 & & \\ & & \ddots & \\ & & & w_{\mathcal{J}} \end{bmatrix} \left( \begin{bmatrix} (g_1^j)^T & & & \\ & (g_2^j)^T & & \\ & & \ddots & \\ & & & (g_{\mathcal{J}}^j)^T \end{bmatrix} \cdot X - \begin{bmatrix} (g_1^k)^T \\ (g_2^k)^T \\ \vdots \\ (g_{\mathcal{J}}^k)^T \end{bmatrix} \right) \right\|_F^2 \quad (5)$$

### 3.3.3. Stiffness regulariser

Allowing each vertex to deform in a locally affine manner leads to an underconstrained optimisation problem with a huge solution space of $\mathcal{J} \cdot 12$ parameters and hence a singular coefficient matrix $A$. As we do not have any shape or deformation priors at hand which could sensibly reduce the number of parameters and hence the size of the parameter search space we instead choose to regularise the problem adding two regularising terms to our cost function.

The first term $\epsilon_s$ controls the overall smoothness of deformation by penalising the weighted difference of the transformation applied to adjacent vertices. This cost is measured as the Frobenius norm $||\cdot||_F$ of the difference of the respective transformation matrices. Adjacency between two vertices a and b is defined as vertex a having a connecting edge to vertex b. The connectivity information is retrieved from the target deformation graph and stored in a neat matrix form using a node-arc incidence matrix $M$ constructed similarly to the formulation in Amberg et al. (2007). Matrix $M^k$ is sparse and of dimension $\mathcal{F} \times \mathcal{K}$ where $\mathcal{F}$ and $\mathcal{K}$ are denoting the number of target faces and nodes respectively. Edge 1 connecting the nodes 2 and 3 would appear as two nonzero entries -1 at $M_{1,2}$ and 1 at $M_{1,3}$. Lowering the stiffness gradually allows more local deformation. Let $I_4$ be a 4 × 4 identity matrix and ⊗ denote the Kronecker product, our stiffness regulariser takes on the form of Eq. (6).

$$\epsilon_s(X^j) := ||(M^k \otimes I)X^j||_F^2 \quad (6)$$

### 3.3.4. Mobility constraints

In every pairwise registration, the source and target graphs are only partially overlapping as we are progressively capturing the torso under rotation exploring unseen geometry with every new frame. The source nodes in a nonoverlap region do not have a correspondence and would otherwise only be driven by the smoothness regulariser. This leads to a 'flattening' of curved surface parts e.g. around the core which does not reflect the actual body shape. Hence, we add mobility constraints that softly enforce vertices without correspondences to stay stationary by quadratically penalising any displacement from their starting location. We implement the mobility constraints by stacking the matrix of the 'parked' nodes $\hat{P}$ on our coefficient matrix $A$ and the matrix $P$ onto the $B$ matrix. $\hat{P}$ and $P$ are similar in definition to the matrices $\hat{G}$ and $G$ in Section 3.3.2 where correspondences $\Theta$ are replaced by self-correspondences $\Theta_s = \{(g_i^j, g_i^j)\}_i$ for all i with $w_i = 0$.

$$\epsilon_m := ||\hat{P}^j \cdot X^j - P^j||_F^2 \quad (7)$$

### 3.3.5. Linear least squares optimisation

Assuming a minimisation of the cost to align graph j to k for fixed stiffness $\alpha$ the total cost $\epsilon$ is the weighted sum of our three terms: the data term $\epsilon_d$, the stiffness term $\epsilon_s$ and the mobility constraints term $\epsilon_m$. The weight for the mobility constraints is fixed to $\beta = 1$. We aim to minimise $\epsilon$ with respect to $X^j$.

$$\min_{X^j} \epsilon(X^j) = \min_{X^j} \epsilon_d(X^j) + \alpha \cdot \epsilon_s(X^j) + \beta \cdot \epsilon_m(X^j) \quad (8)$$

The matrices for the individual cost terms can be conveniently stacked upon as described in Amberg et al. (2007) to yield the linear matrix form below.

$$\epsilon(X^j) = \left\| \begin{bmatrix} \alpha M^k \otimes I \\ W\hat{G}^j \\ \beta \hat{P}^j \end{bmatrix} X^j - \begin{bmatrix} 0 \\ WG^k \\ \beta P^j \end{bmatrix} \right\|_F \quad (9)$$

This system of linear equations in the form $\boldsymbol{AX} = \boldsymbol{B}$ can now be solved for optimal transformations $\boldsymbol{X}^j$ in closed form for a fixed set of correspondences and stiffness.

$$\boldsymbol{X} = (\boldsymbol{A}^T\boldsymbol{A})^{-1} \cdot \boldsymbol{A}^T\boldsymbol{B} \tag{10}$$

As correspondences are implicit and change with every update of $\boldsymbol{X}^j$, optimisation still follows an iterative approach. The iterative process is considered to have converged if the change in $\boldsymbol{X}^j$ defined as $||\boldsymbol{X}^j - \boldsymbol{X}^*||_F$ where superscript $\boldsymbol{X}^*$ equals $\boldsymbol{X}^j$ from the last iteration drops below a threshold of $10^{-4}$ or the maximum number of 10 iterations has been reached.

### 3.4. Spatial propagation

After we estimated an affine transformation for every node $\boldsymbol{g}$ of our deformation graph we wish to apply the deformation to our topologically similar full resolution point clouds. This is achieved by smoothly interpolating the transformation for arbitrary point locations in 3D Euclidean space which we denote by function $X(\cdot)$. The weighted interpolation technique averaging the affine transformation of surrounding nodes is outlined in Sumner et al. (2007) and restated for completeness.

$$\boldsymbol{v}'_i = X(\boldsymbol{v}_i) = \sum_{j=1}^{\mathcal{M}} w_j(\boldsymbol{v}_i)\left[\bar{\boldsymbol{R}}_j(\boldsymbol{v}_i - \boldsymbol{g}_j) + \boldsymbol{g}_j + \bar{\boldsymbol{t}}_j\right] \tag{11}$$

where vertex $\boldsymbol{v}$ is transformed to $\boldsymbol{v}'$. $\bar{\boldsymbol{R}}$ and $\bar{\boldsymbol{t}}$ denote the transformation relative to the node position and $\mathcal{M}$ is the number of nearest node neighbours. Normals are adjusted similarly ignoring the translation part. The precomputed weights for each node radially decay with distance. We take over the weight formulation by Sumner et al. (2007)

$$w_j(\boldsymbol{v}_i) = \frac{1}{\zeta}\left(\frac{1 - ||\boldsymbol{v}_i - \boldsymbol{g}_j||}{d_{max}}\right)^2 \tag{12}$$

$d_{max}$ is the largest vertex-node distance in the set of $\mathcal{M}$ nearest nodes. The weight of node j with respect to vertex i is intuitively decaying with the distance between the node and the vertex. This way we are ensuring local overlapping but limited influence on the final transformation. Finally, weights are normalised to sum to one through use of the normalisation constant $\zeta$.

### 3.5. Model generation

After nonrigid alignment and propagation of the frame and bundle transformations to the full resolution point clouds in a two-step process, each point cloud is fused in a joint model point set $\bigcup_{i=1}^{\mathcal{N}} \tilde{\boldsymbol{V}}^i$. We then run the Moving Least Squares (MLS) algorithm (Alexa et al., 2003) on the resulting model point set. This algorithm fits a low order polynomial to each point over a small spatial neighbourhood reducing noise while maintaining surface detail. The optimal radius depends on the noise level and sampling density. We use a constant search radius of 8 mm. As the algorithm itself is oblivious to the camera location, a subsequent reorientation of flipped 'rogue' normals was necessary. The model size grows linearly with the number of frames being fused leading to oversampling. In a grid-based resampling we reduce point redundancy to a uniform density of 1 mm. To turn the point set into a triangular mesh we apply Poisson reconstruction (Kazhdan and Hoppe, 2013) with a maximum octree depth of 9 and a minimum of 10 samples per point to obtain surface $\boldsymbol{S}$. Excess surface is clipped. For a realistic appearance of the digital torso model, we reintegrate vertexwise colour values via hue interpolation in a radial neighbourhood after meshing. This requires a conversion to and from HSV colour space. Due to the circular nature of the hue component, hue values are sequentially averaged determining the shortest path between two hue values. This avoids colour blending artefacts in regions of stark contrast.

## 4. Experiments

Data was captured using a first generation Microsoft Kinect sensor complying with a clinical acquisition protocol (Section 4.1). We run our method on 9 clinical data sets of 6 patients diagnosed with early breast cancer. Statistics per data set are given in Section 4.2. The presented method is validated in two ways. First, the spread of projected landmark instances is evaluated before and after nonrigid refinement (Section 4.3). Second, breast volume is compared against the rigid-only reconstruction as well as the gold standard (Section 4.4).

### 4.1. Clinical data capture

An acquisition protocol served for a cross-sectional and longitudinal clinical study in the context of an EU-funded project. Kinect and gold standard scans were acquired in distinct rooms over the course of one imaging session by a clinical photographer. Despite using a Kinect in this work, the proposed method is generally camera-agnostic and another RGBD camera could be used interchangeably.

#### 4.1.1. Kinect scan

We employ a first generation Microsoft Kinect sensor which was originally sold as a motion capturing video game interface but found huge appeal in other fields such as 3D surface reconstruction (Zhang, 2012). The Kinect couples medium resolution structured-light depth sensing in the near infrared range with an ordinary RGB camera. Detailed specifications of the Kinect are available online.[1] For Kinect capture, the protocol specifies tripod-mounting the sensor in front of a neutral blue background. The windowless room is evenly lit with diffuse studio lights. The patient stands hands-on-hips at a distance of 0.9 m from the camera and is instructed to perform a 180° self-rotation facing the camera from lateral to lateral while the device captures RGBD data at video frame rate of the patient's torso.

#### 4.1.2. Gold standard scan

A medical-grade 3dMD system[2] was used as the gold standard scanner. It combines four synchronised modular stereo camera units for a 180° torso coverage at a specified geometrical accuracy of < 0.2 mm. Each unit additionally houses a white light pattern projector for active stereophotogrammetry and a uniform flash for texture mapping. Continuous triangular surface meshes of the patients' torsos were generated in single-shot frontal acquisitions. The gold standard models serve as validation data sets in Section 4.4.

### 4.2. Performance

Data was reconstructed offline on a standard PC equipped with an Intel i7 2.8GHz CPU with 32GB of RAM and an Nvidia GeForce GTX 1050 graphics card. Out of the patient cohort, 6 patients with a high distinctiveness of skin features were chosen to faciliate landmark-based validation. For reconstruction, Kinect RGBD sequences were subsampled to about 15 Hz or 1 frame per degree of rotation. The global phase of alignment converged by step 100 for all data sets. A fixed stiffness value $\epsilon_s$ of 20 has been used throughout the experiments to prevent overfitting. Overall timings are reported in Table 1 and relative run times per optimisation step in Fig. 4. We use PCL framework[3] in version 1.8 for point cloud processing and Eigen libraries[4] in version 3.3.4 for sparse matrix arithmetic and linear solvers.

**Table 1**
Overall run time distribution of our method on all clinical cases. Our method runs on a single core making only occasional use of CPU multi-threading or the GPU. The computation time scales near-linearly with the size of the deformation graphs and therefore the degree of downsampling applied internally.

| Patient | # of frames | # of points | Bundle fit | Global fit | MLS | Poisson recon. | Misc. | Sum | # of model vertices |
|---|---|---|---|---|---|---|---|---|---|
| 1 | 152 | 9,386,548 | 22m 44s | 1h 27m 10s | 8m 57s | 4m 03s | 3m 00s | 2h 05m 54s | 317,755 |
| 2 | 131 | 6,818,374 | 23m 15s | 1h 09m 51s | 5m 12s | 3m 12s | 2m 15s | 1h 43m 44s | 275,455 |
| 3 | 151 | 9,128,226 | 19m 04s | 51m 10s | 5m 49s | 2m 42s | 2m 22s | 1h 21m 07s | 214,582 |
| 4 | 157 | 8,730,287 | 16m 14s | 50m 10s | 4m 29s | 3m 37s | 2m 24s | 1h 16m 54s | 297,960 |
| 5 | 117 | 5,150,698 | 15m 44s | 1h 27m 19s | 7m 00s | 3m 04s | 2m 07s | 1h 55m 14s | 244,292 |
| 6 | 101 | 6,561,076 | 16m 50s | 1h 09m 11s | 7m 19s | 3m 21s | 2m 20s | 1h 39m 01s | 278,315 |

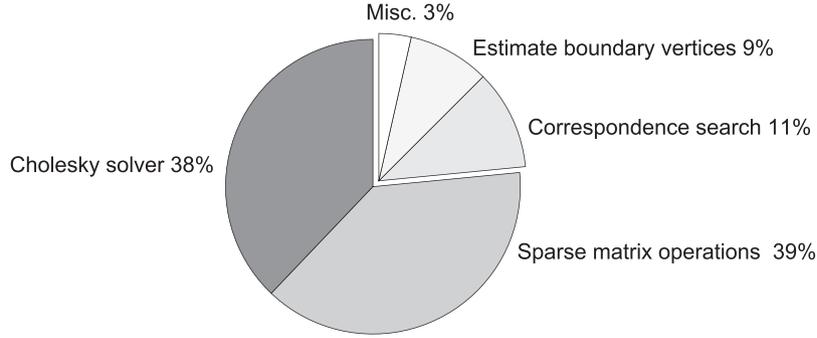

**Fig. 4.** Pie chart with relative timings per optimisation step. The linear solver and corresponding sparse matrix manipulations consume about three-quarter of the processing time.

### 4.3. Landmark-based validation

The challenges in evaluating nonrigid registration algorithms are acknowledged in literature (Tam et al., 2013). In addition to visual assessment we seek a quantitative metric to validate the quality of our proposed nonrigid alignment method in absence of ground truth data for the clinical patient data sets. Surface-to-surface distances from our reconstruction to a gold standard model captured with a high-precision 3D scanner are not suitable for validation due to postural differences between both acquisitions. Instead we devise a landmark validation strategy based on explicit feature correspondences. A prominent feature such as the nipple, beauty spots, skin marks or other salient point becomes a landmark sample instance $l_i^j = (x, y)^T$ where i indexes the landmark with $x \in [1.\mathcal{W}]$ and $y \in [1.\mathcal{H}]$ where $\mathcal{W}$ and $\mathcal{H}$ denote the image resolution. Features are manually picked in a subset $\Omega_i \subseteq \{1, 2, ., \mathcal{N}\}$ of registered 2D colour frames $\tilde{C}^{j \in \Omega_i}$ of the patient's RGBD sequence in which the landmark i could be successfully manually detected. All landmark sample pixel coordinates are backprojected, rigidly transformed into a common coordinate system and subsequently transformed by the distance-weighted interpolated affine transform of neighbouring deformation graph nodes as described in 3.4 which we denote as function $X^j(\cdot)$ for frame j.

$$L_i^j = X^j\left(T_j^{-1} \cdot K_D^{-1} \cdot D(l_i^j)\right) \tag{13}$$

Ideally, after nonrigid refinement all samples of one landmark coincide in the same 3D location with a smaller spread correlating to a better alignment. To obtain the total landmark error score $\epsilon_L$, we compute a per landmark covariance matrix and report the mean Frobenius matrix norm $||\cdot||_F$ over all covariances in Eq. (14). We

```
Require: four contours c
Require: mesh vertices v and faces f
Require: landmarks l
    seed v_seed ← mean of l
    seed face f_seed ← closest face in f to v_seed
    breast faces f_breast ← f_seed
    faces to be checked f_check ← f_seed
    initialise faces already checked f_checked to empty set
    while f_check is not empty do
        clear f_next
        for all f_i in f_check do
            for all edges e_i in f_i do
                if e_i not in c then
                    f_e ← other faces containing e_i
                    if f_e not in f_checked then
                        add f_e to faces to be checked next f_next
                        add f_e to f_breast
                    end if
                end if
            end for
        end for
        add f_check to f_checked
        f_check ← f_next
    end while
    return f_breast
```

**Fig. 5.** Pseudocode formulation of the breast segmentation from contours.

---

[1] https://msdn.microsoft.com/en-us/library/jj131033.aspx
[2] http://www.3dmd.com/static-3dmd_systems/
[3] http://pointclouds.org/
[4] http://eigen.tuxfamily.org

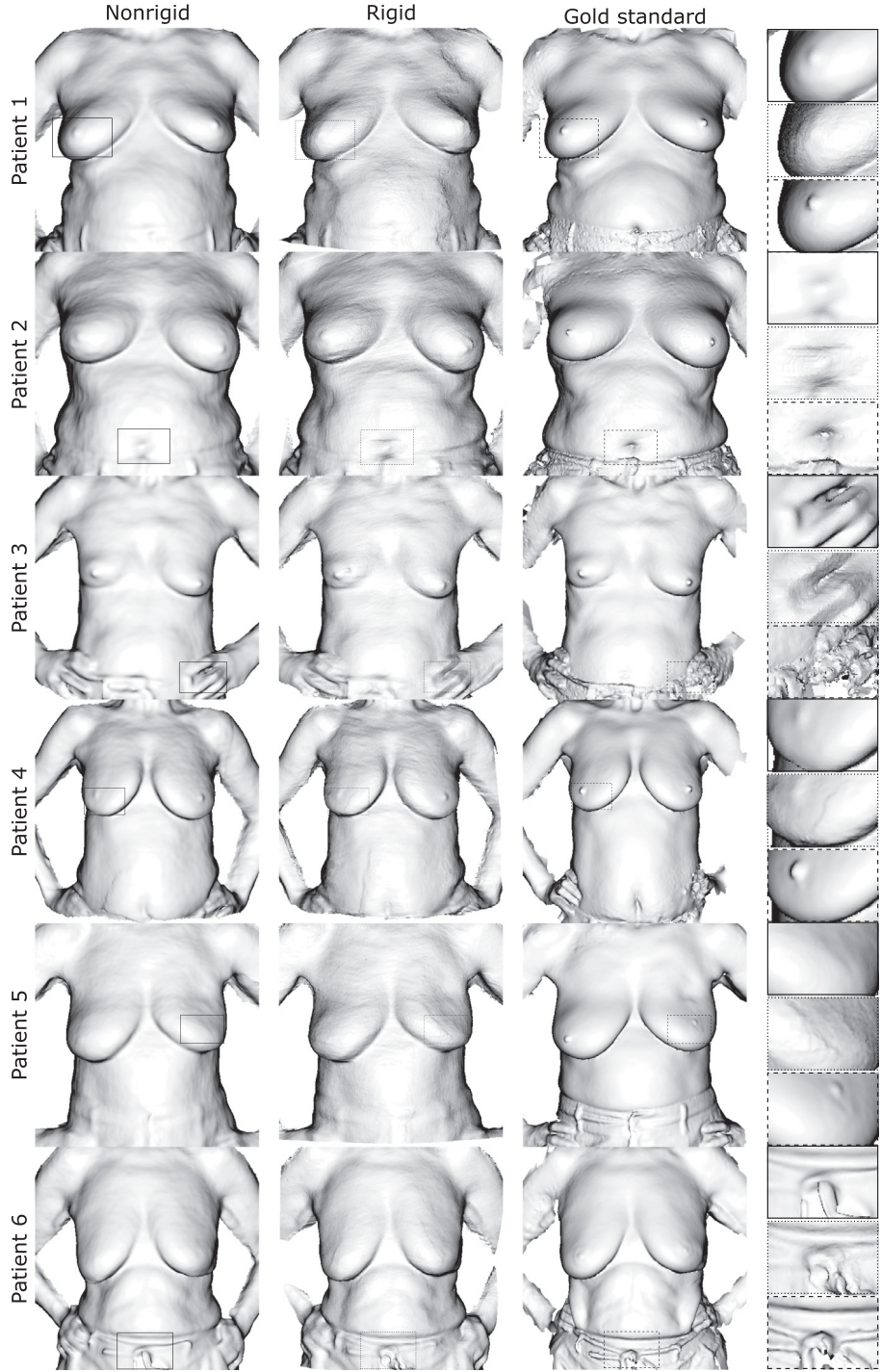

**Fig. 6.** Qualitative results figure comparing Phong-shaded breast surface models from our method (left column), a competitive rigid reconstruction method (middle column) and a high-precision 3D scanner considered gold standard (right column). Accounting for nonrigid deformation our method is able to resolve finer geometric details with less artefacts such as the nipple, belly button or fingers.

let $\mathcal{L}$ denote the number of landmarks.

$$\epsilon_L = \frac{1}{\mathcal{L}} \sum_{i=1}^{\mathcal{L}} ||cov([\mathbf{L}_i^j, ., \mathbf{L}_i^k]^T)||_F \quad with \quad j, k \in \Omega_i \qquad (14)$$

A lower error value $\epsilon_L$ indicates a better nonrigid alignment. Patients' RGB images were visually assessed with respect to a high number, even distribution and small scale of local skin features. Nonetheless, skin features had to be large enough to retain saliency at the given resolution, sharpness and lighting conditions.

Fig. 9 depicts the detected skin features marked as purple dots in frontal and lateral colour images and as blue and green coloured error ellipsoids in 3D space before and after nonrigid alignment. An error ellipsoid represents the multivariate normal distribution fit to all samples of a landmark with elliptic radii of two standard deviations. The number of landmarks and samples per patient are listed in Table 2.

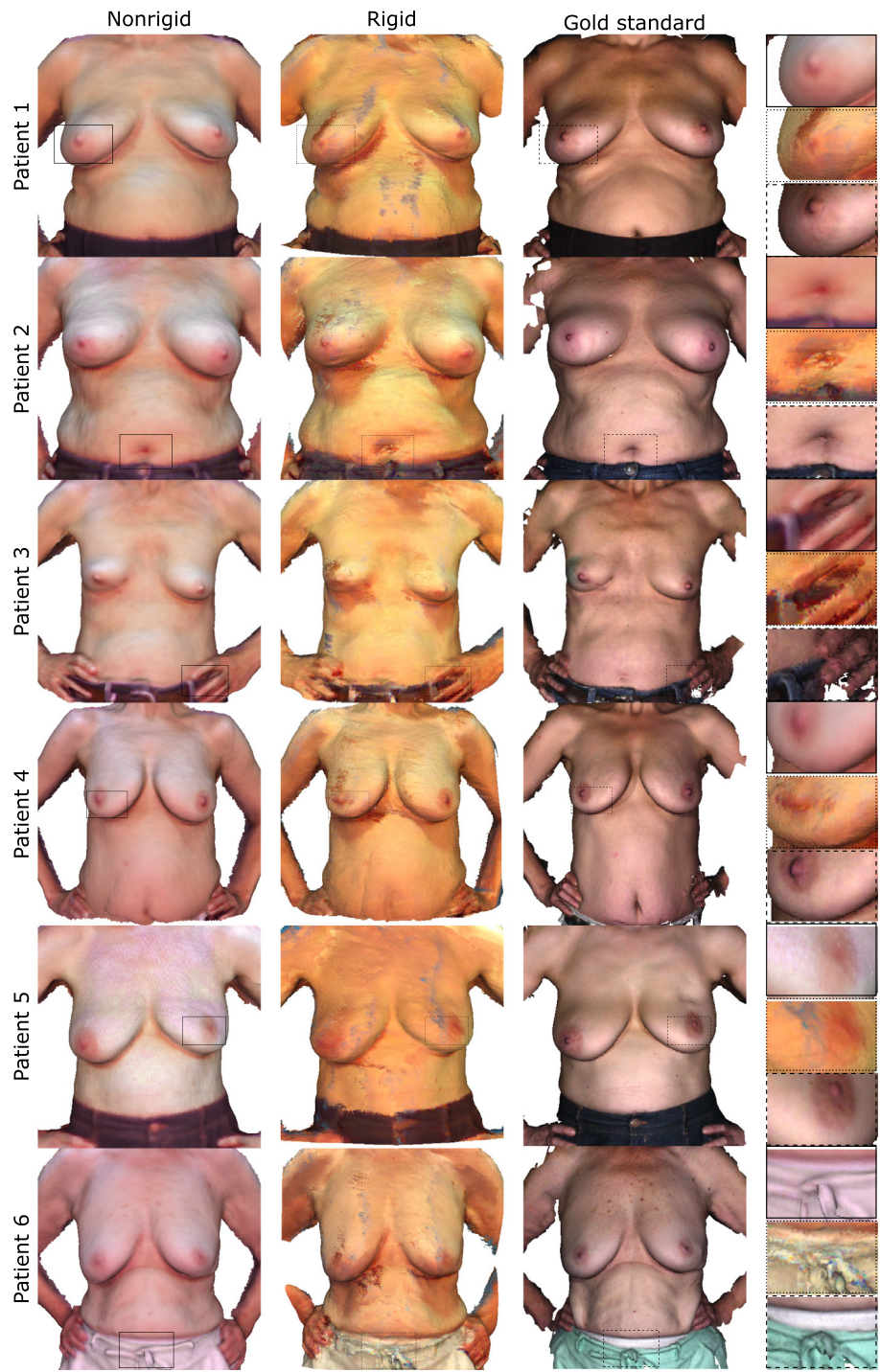

**Fig. 7.** Qualitative results figure comparing breast surface models in a vertexwise colour rendering. The figure follows the same layout as above. Corroborating our previous findings surface texture is also resolved to a higher precision using our method. Beauty spots, moles or the areola are clearly delineated. Background blending and shadowing artefacts are greatly reduced. White balancing leads to an overall more natural colour profile.

**Table 2**
Per patient number of identified landmarks and manually picked landmark samples.

| Patient | 1 | 2 | 3 | 4 | 5 | 6 |
|---|---|---|---|---|---|---|
| # landmarks | 24 | 20 | 19 | 14 | 20 | 21 |
| # samples | 206 | 204 | 154 | 145 | 155 | 227 |

### 4.4. Breast volume assessment

The second part of our validation evaluates breast volume. Breast volume has been studied and reported extensively as a key morphological metric for breast shape since the middle of the last century (Ingleby, 1949). The use of breast volume in our validation is motivated by the assumption that soft tissue is highly deformable yet incompressible. Through breast volumetry we introduce a clinically relevant statistic with respect to breast cosmesis. In clinical practice, breast volume plays a vital role in surgi-cal decision making. The volumetric analysis of the breast complements the landmark-based validation outlined in the previous section in which systematic errors might go undetected. Breast isolation which forgoes volumetry requires estimation of the breast boundaries and chest wall from surface data. This is achieved by defining the breast as a rectangular region of interest along the lines of Catanuto et al. (2018). Volume estimation precede cleanup steps that remove non-manifold edges and vertices as well as non-connected components. Contour corner points of both breasts are algorithmically derived from a set of five manually picked landmarks per torso (sternum, lowest point on inframammary fold and anterior axillary line point bilaterally). Contours are then computed as geodesic paths using Dijkstra's algorithm. The four contours meet at their end points and a grid $G_{Coon}$ is bilinearly blended between them as defined in Eq. (15).

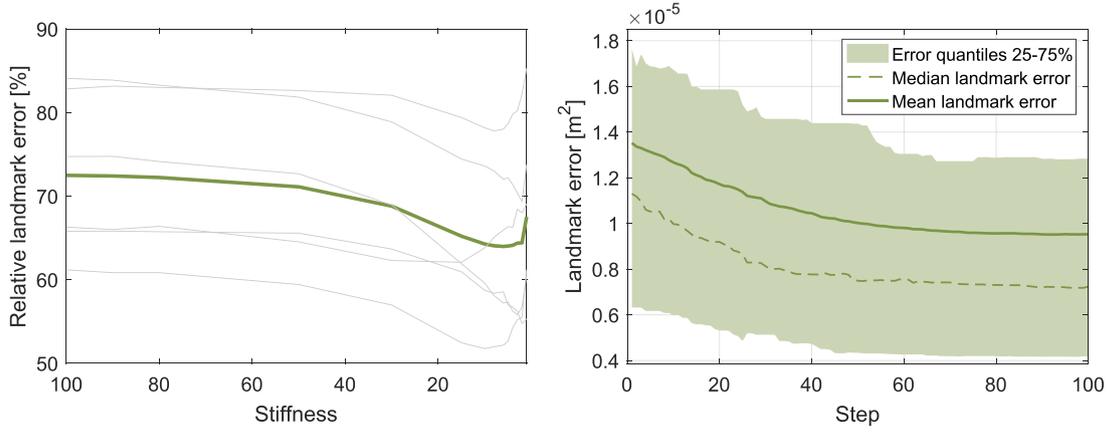

**Fig. 8.** Landmark error versus stiffness chart (left). The green line is the mean over all patients plotted as grey lines. Landmark error is monotonically decreasing with lower stiffness values reaching a minimum in the range of 2 to 15. Quantile plot including all patients shows continuous converging reduction in landmark error over the number of steps (right). (For interpretation of the references to colour in this figure legend, the reader is referred to the web version of this article.)

$$\begin{aligned}
G^x_{Coon}(u,v) = &(1-\hat{u}) \cdot G^x_{Coon}(1,v) + \\
&(1-\hat{v}) \cdot G^x_{Coon}(u,1) + \\
&\hat{u} \cdot G^x_{Coon}(\mathcal{N}_u,v) + \\
&\hat{v} \cdot G^x_{Coon}(u,\mathcal{N}_v) + \\
&(\hat{u}-1) \cdot (1-\hat{v}) \cdot G^x_{Coon}(1,1) + \\
&\hat{u} \cdot \hat{v} \cdot G^x_{Coon}(\mathcal{N}_u,\mathcal{N}_v) + \\
&\hat{u} \cdot (\hat{v}-1) \cdot G^x_{Coon}(\mathcal{N}_u,1) + \\
&\hat{v} \cdot (\hat{u}-1) \cdot G^x_{Coon}(1,\mathcal{N}_v)
\end{aligned} \quad (15)$$

The superscript denotes the coordinate, $u$ and $v$ are the embedded and $\hat{u}$ and $\hat{v}$ are the normalised embedded coordinates or interpolation factors along the contour patch dimensions. The interpolation is equally applicable to the y and z coordinate. Planar as well as Coon's patch chest wall approximations can be found in literature. A Coon's patch has the advantage to be able to model the anatomic breast boundary more flexibly and therefore more accurately. A visualisation of the Coon's torso patches generated from our patient data is shown in the results section in Fig. 10. Overlapping contour segments are deleted prior to patch triangulation. An unequal number of contour points of opposite contours is adjusted beforehand by duplicating points through linearly spacing the contour point indices to the length of the contour comprising more points. This patch closes the breastless torso as seen in Fig. 10. Subsequently, the breast is segmented. A pseudocode formulation is given in Fig. 5. The breast segmentation and torso patch are closed by triangulating all contour edges to a mutual point. This mutual point is found by walking the opposite mean normal direction scaled by the size of the bounding box. The difference volume of both watertight meshes is the sought-after breast volume. Volumes are computed as the polyhedral mass of volume integrals as implemented in vcglib v1.0.1 (Mirtich, 1996).

Breast segmentation including the rear demarcation of the breast on the chest wall is eminently challenging and error-prone using surface data alone. Many breast delineation strategies based on landmark selection have been suggested in literature (Pöhlmann et al., 2017). We argue that our breast segmentation has sufficient repeatability to perform relative volume comparison between methods. We therefore compute the Coefficient of Repeatability (CR) and Coefficient of Variation (CV) which constitute measures of volume precision calculated from the volume differences of the untreated breast for $\mathcal{R} = 3$ patients captured in two imaging sessions pre and postoperatively. The CV indicates the relative magnitude of volume variation with respect to the measurements. Both statistics are proportional to the within-subject variance

$$\sigma^2_w = \sqrt{\sum_{i=1}^{\mathcal{R}} \sigma_i^2 / \mathcal{R}} \quad (16)$$

where $\sigma_i^2$ denotes the variance of the two measurements for breast i.

## 5. Results

The effectiveness of our method is demonstrated qualitatively through renderings of the polygon torso surfaces. We also provide quantitative results from the manual landmark validation as well as the breast volume assessment study. In both cases, we compare against the only competitive breast reconstruction method known to the authors that uses a large number of breast images from a low-cost depth camera for a complete denoised reconstruction which is further referred to as 'rigid' (Lacher et al., 2015). Results are visually superior to those obtained through rigid reconstruction with the latter and fare well against models from a costly commercial 3D scanner as shown in Figs. 6 and 7.

Landmark error is consistently reduced with our method as shown in the bar plots in the right column of Fig. 9. In our 6 patient study, landmark error drops from $1.39 \pm 0.99^{-5}$ m$^2$ with the rigid reconstruction over $1.15 \pm 0.92^{-5}$ m$^2$ using closest point correspondences with our nonrigid reconstruction technique to $0.95 \pm 0.81^{-5}$ m$^2$ switching to shortest distance correspondences. We visualise the reduction in landmark error through shrunken error ellipsoids Fig. 9. Additionally, we show the landmark error behaviour over all patients in dependence of the stiffness and its decrease with the number of nonrigid refinement steps in Fig. 8.

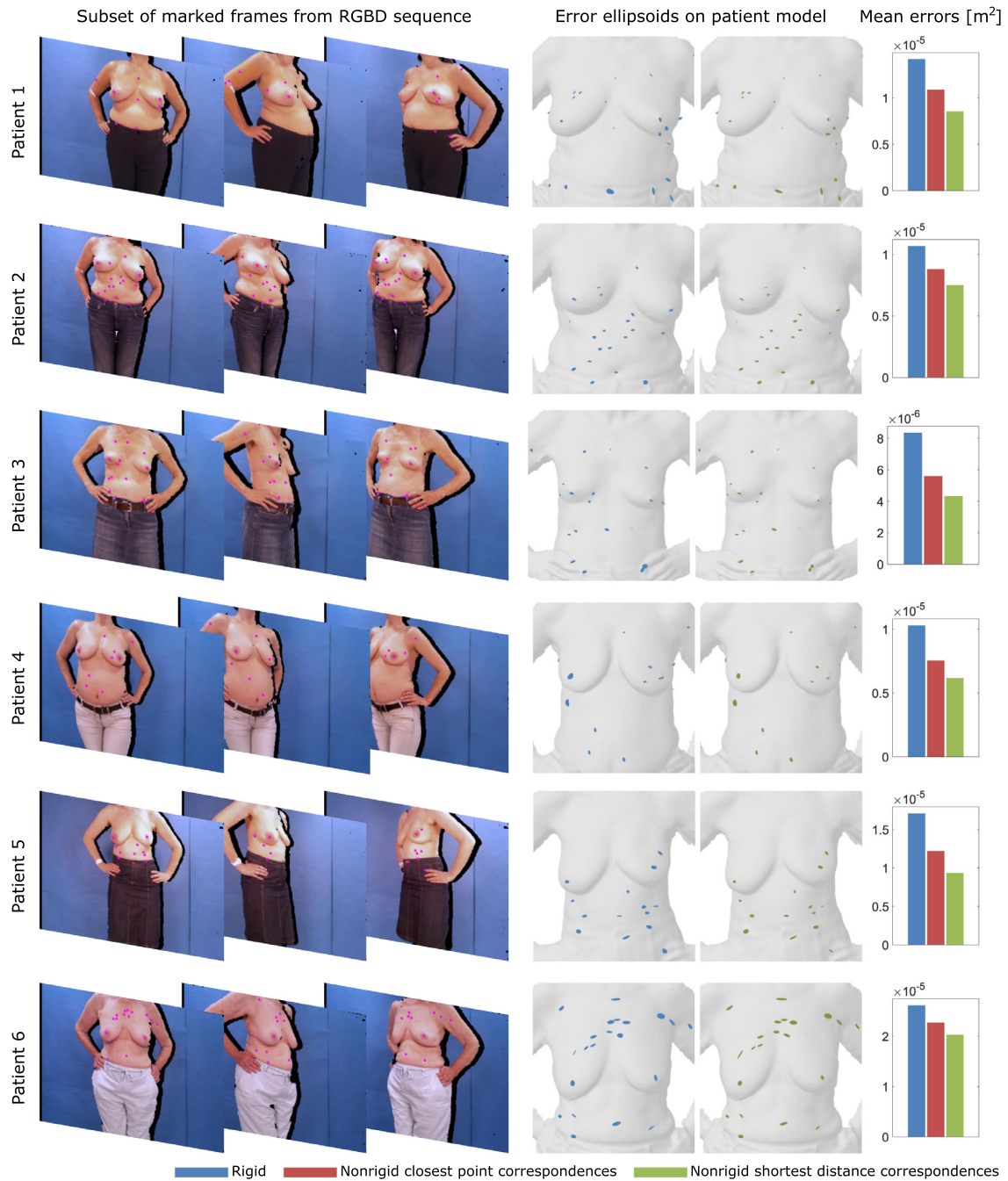

**Fig. 9.** Picked landmark locations are marked by purple dots in exemplary frontal and lateral input colour images (left). Landmark error is visualised as ellipsoids (middle). Samples of each landmark are axis aligned and normal fitted. Ellipsoids axes' length is chosen to represent a 95% confidence interval. The Poisson reconstructed surface mesh (Kazhdan et al., 2006) is overlaid for guidance. Bar plot quantifying landmark error reduction through nonrigid reconstruction using closest point and shortest distance correspondences over sequential rigid reconstruction (right). (For interpretation of the references to colour in this figure legend, the reader is referred to the web version of this article.)

In the breast volumetry part of our validation, we analyse a total of 51 volumes, 17 per method, measured on 11 breasts of 6 patients in 9 imaging sessions. One breast was excluded due to poor lateral coverage in the gold standard scan. A clear linear association between volumes of all methods can be seen in Fig. 11. Owing to the lack of ground truth, volume errors in the gold standard have to be assumed because of which we have chosen to use orthogonal over standard linear regression. Volume differences between methods appear normally distributed in histogram and Q-Q plots (not shown). A $\chi^2$ test for normality failed to reject the null hypothesis that volume differences follow a normal distribution at p=0.86. With a mean and median absolute volume difference to gold standard of 18 and 15 ml over 19.2 and 18.5 ml, results indicate our nonrigid method outperforms the rigid reconstruction in terms of breast volume accuracy by a small margin. The Bland-Altman plots in the right column of Fig. 11 also show that volume differences do not seem to increase in absolute values with breast size. An examination of volume difference outlier revealed misestimations in the chest wall approximation caused by subtle changes in the patient's posture between modalities, especially in the presence of saggy breasts with a partly occluded inframammary fold. These samples might therefore not reliably reflect the actual volume difference. Furthermore, both the proposed as well as the rigid method suggest a minute volume bias of 4 and 3 ml, respectively. Repeatability of the non-rigid method is, with 5.2%, marginally worse than the gold standard with 5.1% and more than twice as high compared to the rigid method with 2.2%, as per CV in Table 3.

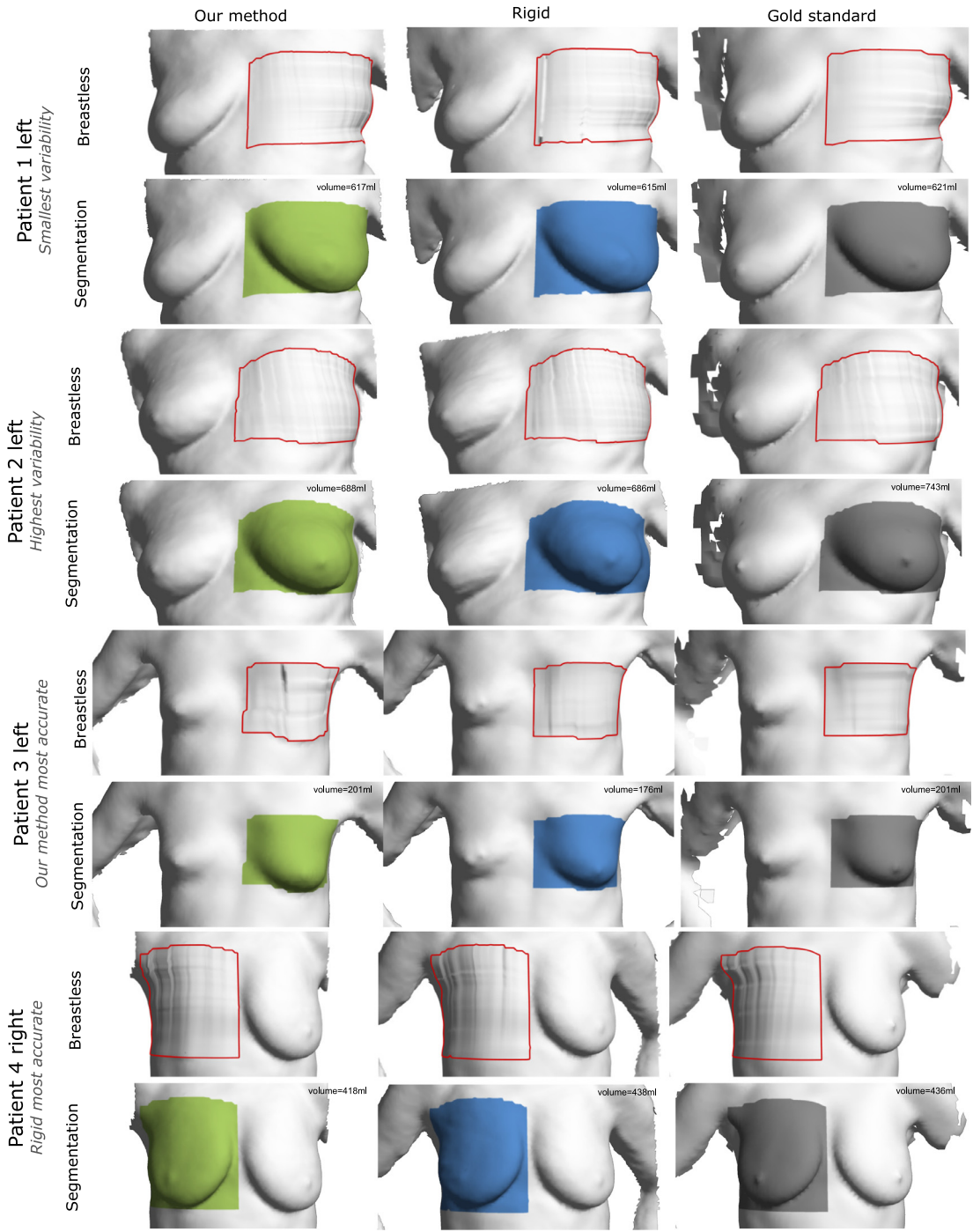

**Fig. 10.** Qualitative results in the breast volume evaluation. Methods are compared columnwise. Exemplary cases are shown. Top to bottom: The least and most variable volume estimation among all three methods, the most accurate volume measurement as compared to ground truth using the proposed method and its rigid predecessor. Per case the approximated breastless torso with contoured breast region and the breast segmentation are shown on top of each other.

## 6. Discussion

Low-cost lightweight 3D scanning of a human torso always goes hand in hand with some inherent degree of complex deformation.

**Table 3**
Coefficients of repeatability and variation calculated using the healthy breast volume differences from two imaging sessions in a small subset of three longitudinal patients. Postural and body mass changes, discretisation and manual landmarking contribute to a higher variability in volume estimates.

|  | Nonrigid | Rigid | Gold standard |
|---|---|---|---|
| CR [ml] | 63.9 | 34.8 | 57.6 |
| CV [%] | 5.2 | 2.2 | 5.1 |

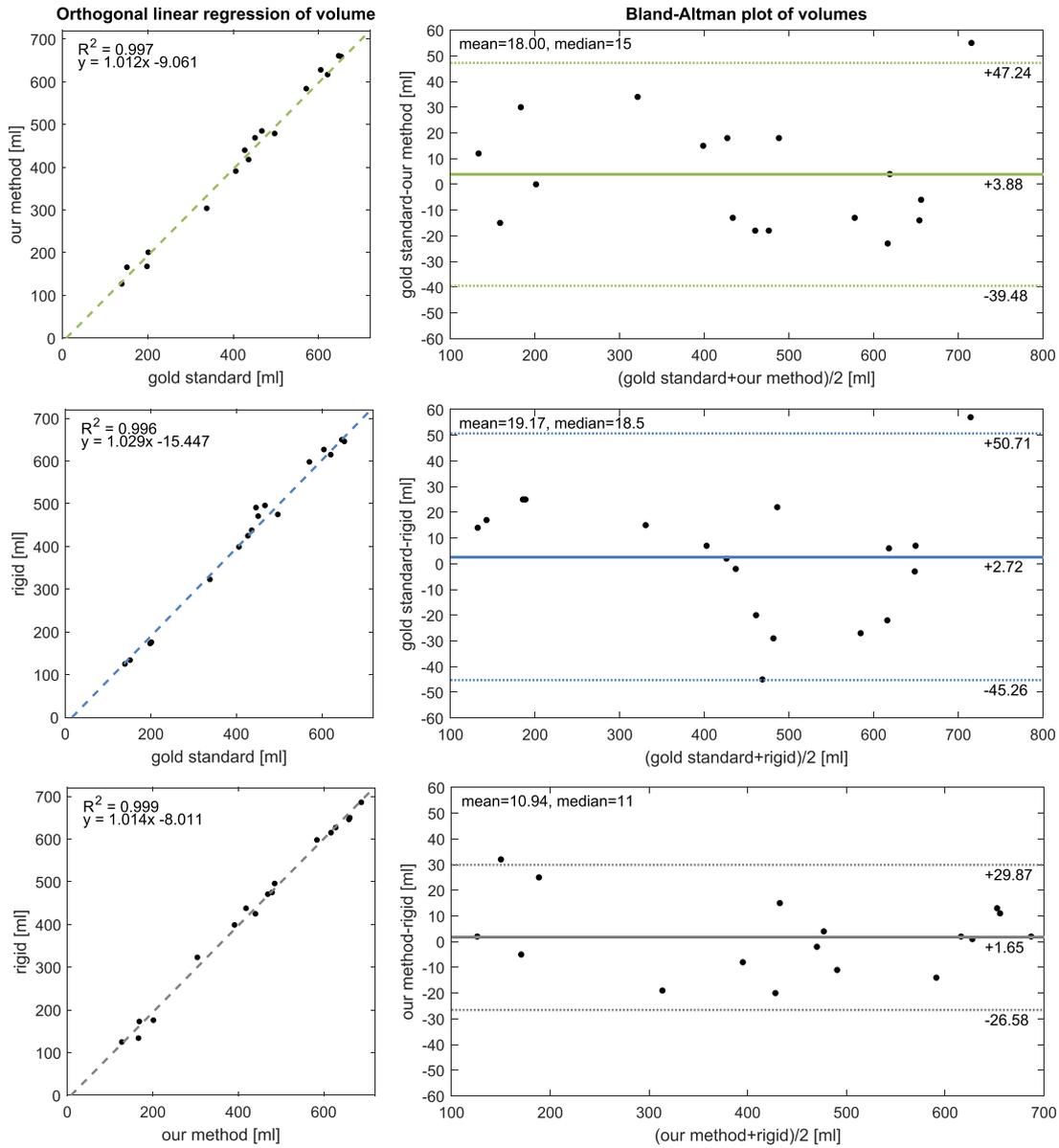

**Fig. 11.** Quantitative results in breast volume estimation. The left column shows linear regression and the right column Bland–Altman plots for the estimated breast volume data. Each row analyses two out of three methods against each other. The analysis suggests a narrow gain in accuracy and precision of the proposed over the rigid method.

This is due to the fact that the image data from various viewpoints necessary for a complete hole-free 3D model has to be captured consecutively. Between these scans a certain amount of motion is hard to avoid. Soft tissue deformation is increased in particular in the domain of breast scanning when compared to faces or other body parts with an underlying bone structure. The low price and portability of such cameras preclude a setup with multiple calibrated camera pods capable of single shot acquisition. Moreover, noise level tends to be higher with cheaper hardware suggesting the fusion of a larger number of images for noise suppression. This should be reflected in the reconstruction approach by appropriately modelling the interframe transformation. Moving from rigid to nonrigid registration is not trivial as the problem of nonrigid registration is ill-posed. The mainly homogeneous skin texture of the breast additionally hampers the use of photometric consistency to guide the alignment. Bearing these challenges in mind we developed an algorithmic solution to human breast and torso surface reconstruction moving on from purely rigid reconstruction integrating a nonrigid correction for involuntary body motion. This globally consistent nonrigid ICP-based refinement is not restricted to a learnt shape or motion model. Instead we assume deformation between frames is small to be able to generically correct any deformation without explicitly constraining it other than requiring it to be smooth.

Due to the fact that the torso is featureless for the most part, landmarks are sparse and not evenly distributed across the surface leaving regions in which the quality of registration cannot be quantified. Specularities causing an oversaturation of RGB values are aggravating this circumstance. We also acknowledge that landmark sample proximity in 3D space, that is the closeness of the backprojection of matching features in multiple frames, is a necessary but not sufficient condition for the alignment to be accurate. An intuitive counter example is the case of a degenerated reconstruction where all points are mapped onto one location with a landmark error of 0. Yet, regularisation successfully thwarts such scenarios and we deem the landmark error score to be a good predictor of registration quality.

According to literature, breast volume differences of up to 50 ml are regarded as satisfactory for clinical application (Henseler et al., 2014; Sigurdson and Kirkland, 2006). Compared to gold standard,

24  R.M. Lacher, F. Vasconcelos and N.R. Williams et al. / Medical Image Analysis 53 (2019) 11–2516 out of 17 breasts reconstructed with the proposed method were measured to be within this error margin of 50 ml, 14 of which were accurate to 20 ml. A similarity in the limits of agreement between the rigid and nonrigid method is to be noted in Fig. 11. This likely happens since both methods feed on the same Kinect input data whereas the gold standard model is acquired under postural change. Although volume estimates should be unaffected assuming breast tissue incompressibility, the fuzziness of breast segmentation may lead to similar variability characteristics. Various sources of uncertainty in the breast segmentation limit the interpretability of the volume evaluation including manual landmarking, mesh discretisation and posture variation. This is exacerbated by a lack of consensus on breast segmentation and volume calculation (Choppin et al., 2016).

We have further found that the stiffness parameter $\epsilon_s$ correlates to breast volume bias which might possibly be attributed to overfitting. It is generally known that a stronger regularisation encourages isometry. This is particularly relevant in the absence of isometry-preserving constraints. Given the variability of the volume differences, the finding of a volume bias of 4 ml shall however be discounted. Also, while our algorithm shows an increase in reconstruction quality using a single stiffness over the whole surface, the amount of surface detail and curvature varies strongly and a locally adaptive smoothness may be advisable. This would come at the cost of model complexity.

Repeatability of breast surface reconstruction is assessed through repeated breast volume measurements on longitudinal data. The coefficients in Table 3 suggest that the rigid reconstruction is the most precise whereas the proposed nonrigid approach is on par with the gold standard. For comparison, the repeatability coefficients obtained in a previous study using a Kinect camera for breast volume measurement were larger by a factor of 2–4 depending on patient pose (Pöhlmann et al., 2017). Although comparing favourably to the latter study with respect to repeatability, results have to be interpreted with caution due to the small sample size and the exclusion of patients with overly ptotic breasts that do not permit occlusion-free capture. The discrepancy in repeatability might also be explained by the considerably larger parameter space associated with the nonrigid deformations. More practically, volume precision estimates such as the CR and CV are also affected across methods by substantial weight gain or loss between imaging sessions. A full statistical analysis on more data would be necessary for reliable conclusions about the repeatability of breast volume measurements using the proposed method.

In terms of performance, note that although real-time is not a requirement the execution times in the range of 1–2 h listed in Table 1 might impede our method's utility in practice. However, our code is highly unoptimised and mass parallelism on the graphics card could be exploited to a higher degree for run-time acceleration.

## 7. Conclusion

Breast cancer is a complex global health problem. With ever rising incidence numbers and longer survival rates along evidence of a correlation between poor aesthetic outcome and quality of life, treatment planning and objective outcome evaluation through the use of 3D breast surface scanning is an important research path. In the dense reconstruction of 3D breast surfaces from low-cost clinical RGBD video, models suffer from oversmoothing and artefacts due to postural sway during lengthy data acquisition. We successfully drop the rigid scene assumption demonstrating more geometric detail and better texture mapping quality than any previous approach. Quantitatively, we are able to show a consistently better alignment for all patients using the proposed landmark error metric formulation and an accuracy within 20 ml in more than three-quarters of breast volume measurements compared to gold standard. In achieving a high reconstruction quality through mitigation of postural sway we believe to overcome a major obstacle towards routine clinical use of low-cost 3D breast surface modelling.

## Acknowledgement

The authors gratefully acknowledge financial support from the EC FP7 PICTURE project (FP7-ICT-2011-9, 600948), the EPSRC (EP/N013220/1, EP/N022750/1, EP/N027078/1, NS/A000027/1) and the Wellcome/EPSRC Centre for Interventional and Surgical Sciences (WEISS) at UCL (203145Z/16/Z).## References

Alexa, M., Behr, J., Cohen-Or, D., Fleishman, S., Levin, D., Silva, C.T., 2003. Computing and rendering point set surfaces. IEEE Trans. Vis. Comput. Graph. 9 (1), 3–15.
Amberg, B., Romdhani, S., Vetter, T., 2007. Optimal step nonrigid icp algorithms for surface registration. In: 2007 IEEE Conference on Computer Vision and Pattern Recognition. IEEE, pp. 1–8.
Besl, P.J., McKay, N.D., et al., 1992. A method for registration of 3-d shapes. IEEE Trans. Pattern Anal. Mach. Intell. 14 (2), 239–256.
Bogo, F., Black, M.J., Loper, M., Romero, J., 2015. Detailed full-body reconstructions of moving people from monocular rgb-d sequences. In: Proceedings of the IEEE International Conference on Computer Vision, pp. 2300–2308.
Cardoso, M.J., Cardoso, J.S., Vrieling, C., Macmillan, D., Rainsbury, D., Heil, J., Hau, E., Keshtgar, M., 2012. Recommendations for the aesthetic evaluation of breast cancer conservative treatment. Breast Cancer Res. Treat. 135 (3), 629–637.
Cardoso, M.J., Oliveira, H., Cardoso, J., 2014. Assessing cosmetic results after breast conserving surgery. J. Surg. Oncol. 110 (1), 37–44.
Catanuto, G., Taher, W., Rocco, N., Catalano, F., Allegra, D., Milotta, F.L.M., Stanco, F., Gallo, G., Nava, M.B., 2018. Breast shape analysis with curvature estimates and principal component analysis for cosmetic and reconstructive breast surgery. Aesthetic Surg.J..
Chae, M.P., Rozen, W.M., Spychal, R.T., Hunter-Smith, D.J., 2016. Breast volumetric analysis for aesthetic planning in breast reconstruction: a literature review of techniques. Gland Surg. 5 (2), 212.
Choppin, S., Wheat, J., Gee, M., Goyal, A., 2016. The accuracy of breast volume measurement methods: a systematic review. The Breast 28, 121–129.
Costa, P., Monteiro, J.P., Zolfagharnasab, H., Oliveira, H.P., 2014. Tessellation-based coarse registration method for 3d reconstruction of the female torso. In: Bioinformatics and Biomedicine (BIBM), 2014 IEEE International Conference on. IEEE, pp. 301–306.
Cui, Y., Chang, W., Nöll, T., Stricker, D., 2012. Kinectavatar: fully automatic body capture using a single kinect. In: Asian Conference on Computer Vision. Springer, pp. 133–147.
Curless, B., Levoy, M., 1996. A volumetric method for building complex models from range images. In: Proceedings of the 23rd Annual Conference on Computer Graphics and Interactive Techniques. ACM, pp. 303–312.
Dai, A., Nießner, M., Zollhöfer, M., Izadi, S., Theobalt, C., 2017. Bundlefusion: real-time globally consistent 3d reconstruction using on-the-fly surface reintegration. ACM Trans. Graph. (TOG) 36 (3), 24.
De Angelis, R., Sant, M., Coleman, M.P., Francisci, S., Baili, P., Pierannunzio, D., Trama, A., Visser, O., Brenner, H., Ardanaz, E., et al., 2014. Cancer survival in europe 1999–2007 by country and age: results of eurocare-5a population-based study. Lancet Oncol. 15 (1), 23–34.
Dou, M., Khamis, S., Degtyarev, Y., Davidson, P., Fanello, S.R., Kowdle, A., Escolano, S.O., Rhemann, C., Kim, D., Taylor, J., et al., 2016. Fusion4d: real-time performance capture of challenging scenes. ACM Trans. Graph. (TOG) 35 (4), 114.
Eiben, B., Lacher, R., Vavourakis, V., Hipwell, J.H., Stoyanov, D., Williams, N.R., Sabczynski, J., Bülow, T., Kutra, D., Meetz, K., et al., 2016. Breast conserving surgery outcome prediction: apatient-specific, integrated multi-modal imaging and mechano-biological modelling framework. In: International Workshop on Digital Mammography. Springer, pp. 274–281.
Fisher, B., Anderson, S., Bryant, J., Margolese, R.G., Deutsch, M., Fisher, E.R., Jeong, J.-H., Wolmark, N., 2002. Twenty-year follow-up of a randomized trial comparing total mastectomy, lumpectomy, and lumpectomy plus irradiation for the treatment of invasive breast cancer. N. Engl. J. Med. 347 (16), 1233–1241.
Fitzmaurice, C., Dicker, D., Pain, A., Hamavid, H., Moradi-Lakeh, M., MacIntyre, M.F., Allen, C., Hansen, G., Woodbrook, R., Wolfe, C., et al., 2015. The global burden of cancer 2013. JAMA Oncol. 1 (4), 505–527.
Glocker, B., Shotton, J., Criminisi, A., Izadi, S., 2015. Real-time rgb-d camera relocalization via randomized ferns for keyframe encoding. IEEE Trans. Vis. Comput. Graph. 21 (5), 571–583.
Hartley, R., Zisserman, A., 2000. Multiple View Geometry in Computer Vision, Second Edition, pp. 153–177.
Hau, E., Browne, L., Capp, A., Delaney, G.P., Fox, C., Kearsley, J.H., Millar, E., Nasser, E.H., Papadatos, G., Graham, P.H., 2013. The impact of breast cosmetic and functional outcomes on quality of life: long-term results from the st. george and wollongong randomized breast boost trial. Breast Cancer Res. Treat. 139 (1), 115–123.